\documentclass{article}

\PassOptionsToPackage{square,numbers}{natbib}
\usepackage[final]{neurips_2023}

\usepackage[utf8]{inputenc} % allow utf-8 input
\usepackage[T1]{fontenc}    % use 8-bit T1 fonts
\usepackage{hyperref}       % hyperlinks
\usepackage{url}            % simple URL typesetting
\usepackage{booktabs}       % professional-quality tables
\usepackage{amsfonts}       % blackboard math symbols
\usepackage{nicefrac}       % compact symbols for 1/2, etc.
\usepackage{microtype}      % microtypography
\usepackage{xcolor}         % colors
\usepackage{graphicx}
\usepackage{svg}
\usepackage{multirow}
\usepackage{rotating}% Math package
\usepackage{amsmath}
\usepackage{wrapfig}

\DeclareMathOperator*{\argmin}{arg\,min}

\newcommand{\keypoint}[1]{\noindent\textbf{#1}\quad}
\title{Is Scaling Learned Optimizers Worth It? Evaluating The Value of VeLO's 4000 TPU Months}

\author{%
  Fady Rezk\\
  School of Informatics\\
  University of Edinburgh\\
  \texttt{F.R.G.Rezk@sms.ed.ac.uk}
  \And
   Antreas Antoniou\\
   School of Informatics\\
    University of Edinburgh\\
    \texttt{a.antoniou@ed.ac.uk}
    \And
    Henry Gouk\\
    School of Informatics\\
    University of Edinburgh\\
    \texttt{henry.gouk@ed.ac.uk}
   \And
   Timothy M. Hospedales\\
    University of Edinburgh\\
    Samsung AI Research, Cambridge\\
    \texttt{t.hospedales@ed.ac.uk}\\
}

\begin{document}

\maketitle

\begin{abstract}
  We analyze VeLO (versatile learned optimizer \cite{metz2022VeLO}), the largest scale attempt to train a general purpose ``foundational'' optimizer to date. VeLO was trained on thousands of machine learning tasks using over 4000 TPU months with the goal of producing an optimizer capable of generalizing to new problems while being hyperparameter free, and outperforming industry standards such as Adam. We independently evaluate VeLO on the MLCommons optimizer benchmark suite. We find that, contrary to initial claims: (1) VeLO has a critical hyperparameter that needs problem-specific tuning, (2) VeLO does not necessarily outperform competitors in quality of solution found, and (3) VeLO is not faster than competing optimizers at reducing the training loss. These observations call into question VeLO's generality and the value of the investment in training it.
\end{abstract}

\section{Introduction}
Meta-learning, or learning to learn, refers to the appealing vision of learning the learning algorithm itself, similarly to how deep learning replaced the tradition of handcrafted feature engineering \cite{9428530}. Meta-learning has found compelling applications in various facets of AI. In particular, one notable application of meta-learning is to learn improved optimization strategies \cite{10.5555/3157382.3157543} that provide better or faster optimization than hand-crafted optimizers \cite{chen2022learning}. After initial successes in relatively small scale problems, researchers have recently focused on scaling learned optimizers \cite{10.1007/978-3-031-20050-2_23,metz2022VeLO}.

A noteworthy example is VeLO \cite{metz2022VeLO}. Trained on a huge array of tasks with over 4000 TPU months,  it aspires to be a `foundational' optimizer capable of solving  any new problems more rapidly than hand-designed optimizers such as Adam. VeLO claimed multiple remarkable abilities% to justify its meta-training cost
, such as being at least $4\times$ faster than Adam on 50\% of tasks in the VeLOdrome suite. If true, VeLO would eventually pay for its up-front training cost by accelerating learning across the community. Nevertheless, evaluating optimizers -- especially learned optimizers -- is itself a very difficult problem with multiple facets including iteration and time-efficiency, quality and generalisation of minima discovered, hyperparameter sensitivity \cite{dahl2023benchmarking,chen2022learning,choi2020on}, and generalisation of the learned optimiser itself.

In this work, we critically analyse VeLO's performance to understand if it is as effective as claimed. Our evaluation casts doubt on its claimed efficacy, and whether scaling-up training is the silver bullet for optimizer learning that it has been in other areas of AI.
% Our evaluation casts doubt over the value of scaling learned optimizers. We find that learned optimizer might be one problem that can not be solved by the spending of more compute on model training with growing data sizes. 
%In section \ref{sec:prelim}, we introduce the problem formulation of learning to optimize (LO). This is followed by an overview of LO research, and work done in benchmarking optimizers in \ref{sec:back}. Finally, the benchmark design, experiments and results are presented in sections \ref{sec:exp} and \ref{sec:results}.
Our contributions are:
(1) Validation of VeLO: We conduct a rigorous, independent evaluation of VeLO's performance using an extended analysis based on the \hyperlink{https://MLCommons.org/en/groups/research-algorithms/}{MLCommons} benchmark.%, providing a critical assessment of the optimizer's purported advantages.
(2) Claims Reassessment: Our empirical results challenge several key claims made in the original VeLO paper, specifically that of being hyperparameter-free, outperforming baselines in minimizing training objectives, and offering optimization speedups. %outperforming HPO when generalizing to validation sets and offering optimization speedups
(3) Introduction of Explicit Metrics: We introduce a set of carefully selected metrics that directly align with the fundamental objectives an optimizer should fulfil. These metrics serve as a standardized framework for comparing VeLO against other optimizers.

\section{Preliminaries}\label{sec:prelim}
Let $f_{\theta}$ be a function parameterized by $\theta$ where $\theta$ is defined over some domain $\theta\in\Theta$. We refer to $f_\theta$ as the \emph{optimizee}; the function being optimized. Performance of $f_\theta$ on a task $\mathcal{T}_i$ sampled from a distribution of tasks $p(\mathcal{T})$ can be measured by a loss function $L_i(f_\theta, \mathcal{T}_i)$. The goal of learning is to find the minimizer $\theta^*=\argmin_{\theta\in\Theta}L_i(f_\theta,\mathcal{T}_i)$. Gradient descent minimizes the loss function by producing a sequence of updates in the form:
\begin{equation}
    \theta_{t+1}=\theta_t-\alpha_t\nabla_\theta L_i(f_\theta,\mathcal{T}_i)
\end{equation}
Learning to optimize strategies reformulate gradient descent as 
\begin{equation}
\theta_{t+1}=\theta_t+g(L_i(f_\theta,\mathcal{T}_i)),\label{eq:learnedOptimizer}
\end{equation}
%$\theta_{t+1}=\theta_t+g(L_i(f_\theta,\mathcal{T}_i))$, 
which recovers standard gradient descent when $g(\cdot)$ is a simple scaling $g(L_i(f_\theta,\mathcal{T}_i))=-\alpha_t\nabla_\theta L_i(f_\theta,\mathcal{T}_i)$. These approaches assume that performance can be improved by paramaterizing the function $g$ with some learnable parameters $\lambda$, e.g., defining a small MLP. %By iteratively applying the functional $g$ to minimize the loss at hand, a bi-level optimizaiton problem is produced.

Learning-to-optimize is usually formulated as a bi-level optimization problem where the goal is to the learn \emph{optimizer} $g$ so that the \emph{optimizee} $f$ achieves low loss on some task distribution after learning.  More specifically: 
%To illustrate how an optimization algorithm can be learned, parameterize the optimizer function $g$ by $\lambda$, i.e: an MLP. Then, the goal is to achieve low loss on some task distribution. Then, the bi-level problem can be written as:
\begin{align}
    \lambda^*=\argmin_\lambda\sum_{i=1}^M\mathcal{L}(\theta^*_i(\lambda),\lambda,\mathcal{T}_i)\label{eq:outer} \\
    \text{s.t. } \theta^*_i(\lambda)=\argmin_\theta L_i(\theta,\lambda,\mathcal{T}_i)\label{eq:inner}
\end{align}
where Eq.~\ref{eq:inner} is solved with the learnable optimizer Eq.~\ref{eq:learnedOptimizer}, and the optimizer learning objective is in Eq.~\ref{eq:outer}. 
Compactly, the gradient of the loss, after $t$ steps, on a given sampled task would then be \cite{metz2022gradients}:
\begin{equation}
    \frac{d L_t}{d\lambda}=\frac{\partial L_t}{\partial\lambda}+\sum_{k=1}^T\frac{\partial L_t}{\partial\theta_t}\left(\prod_{i=k}^T\frac{\partial\theta_i}{\partial\theta_{i-1}}\right)\frac{\partial\theta_k}{\partial\lambda},
\end{equation}
which allows the optimizer to be learned with gradient descent. 

\section{Background and Motivation}\label{sec:back}
\keypoint{Learning Optimizers}
Searching for simple and symbolic update rules for training neural networks dates back to the 90's \citep{Runarsson,Bengio}. More recently, \cite{10.5555/3157382.3157543} parameterized the optimization algorithm as an LSTM which acts coordinate-wise on the inner-loop problem. Various work has since explored the design space of learning optimizers. The space spans a) the parameterization of the learned optimizer including it's IO representation, b) the meta-training task distribution, c) meta-optimizers (optimizer used to update the learned optimizer) and d) the outer-loop objective function for estimating the learned optimizer performance.

Parameterizations included LSTMs \cite{10.5555/3157382.3157543,10.5555/3305890.3305913}, hierarchical RNNs \cite{10.5555/3305890.3306069,10.1007/978-3-031-20050-2_23}, MLPs \cite{pmlr-v97-metz19a}, transformers \cite{gartner2023transformerbased} and hyper-networks \cite{metz2022VeLO}. Tree structured search spaces \cite{chen2023symbolic}, domain-specific languages \cite{pmlr-v70-bello17a} and evolutionary strategies \cite{lange2023discovering} were also explored. Search spaces and black-box parameterizations can be learned using various techniques such as gradient-based meta-learning \cite{10.5555/3157382.3157543}, or evolutionary strategies \cite{chen2023symbolic}. Meta-loss functions also vary between inner-loop training \cite{10.5555/3157382.3157543,10.5555/3305890.3305913}, validation loss \cite{pmlr-v97-metz19a,xiong2020improved}, or more complex objectives that measure resource-efficiency \cite{10.1007/978-3-030-58545-7_29} and speed \cite{yang2023mlo}.

\keypoint{Benchmarking optimizers}  Benchmarking optimizers -- especially in deep learning -- is extremely challenging, as there are many facets to optimizer quality including iterations and clock-time to convergence, quality of the solution found in non-convex problems, generalisation of the final solution to a validation or testing set, hyperparameter sensitivity, consistency of performance across different workloads, etc \cite{pmlr-v119-sivaprasad20a,choi2020on,pmlr-v139-schmidt21a}. As discussed in \cite{dahl2023benchmarking}, this is the reason behind multiple apparently contradictory claims in the literature, and the lack of consensus on benchmarks and metrics compared to other areas of machine learning and AI. All this makes it challenging to compare optimizers as they may excel at one facet while falling down in another.

%For practical purposes, when benchmarking optimizers, it is important to (a) account for the cost of hyperparameter optimization (HPO) versus absolute performance \cite{pmlr-v119-sivaprasad20a} while (b) factoring in how adaptive optimizers can approximate other ones \cite{choi2020on} for practical purposes. Optimizer cost is also of practical importance so benchmarks (c) must consider optimizers speed \cite{dahl2023benchmarking}. Moreover, ranking optimizers is challenging as it is known that no optimizer dominates when algorithms are evaluated across diverse workloads. To make matters worse, rankings are sensitive to workload details, hyperparameters search spaces, and tuning goals \cite{dahl2023benchmarking}. Besides, precisely defining and measuring training speed is an ill-defined problem. On the other hand, the self-tuning regime, i.e: optimizers with default hyperparameters, approximately ranks optimizers similarily to HPO setting \cite{pmlr-v139-schmidt21a} performance-wise. In summary, benchmarking optimizers must control for a variety of factors of variation under multiple regimes (self-tuning vs HPO). The main evaluation objectives influenced by the factors of variations are cost/efficiency and absolute performance.

For the reasons discussed above, apples-to-oranges comparisons are common in the literature, and can lead to misleading conclusions. For example, comparing optimizer performance without controlling hyperparameter tuning or HPO objective \cite{pmlr-v119-sivaprasad20a}. This has led to a few attempts to establish common evaluation frameworks for optimizers, notably MLCommons \cite{dahl2023benchmarking}, which can control for HPO.

\keypoint{Benchmarking learned optimizers} 
This challenge of optimizer evaluation is further exacerbated when considering benchmarking of learned optimizers, as the cost of optimizer learning, and the robustness of the learned optimizer to diverse and out-of-distribution tasks open up additional important criteria. 
As optimizer learning is a costly process, most learned optimizers justify themselves with amortization arguments: The idea that the up-front cost of optimizer learning can be paid off by the learned optimizer's improved solution to multiple subsequent tasks. However, the learned optimizer needs to be applied on new tasks for this justification to hold, as good solutions to the training tasks have already been found during optimizer training. Thus, the practical value of a learned optimizer is intrinsically intertwined with both its efficacy and how well it generalizes  to new tasks. All this makes fairly benchmarking learned optimizers even harder than handcrafted optimizers. 

The VeLO optimizer \cite{metz2022VeLO} aspired to achieve both efficient optimization and cross-task generalization by large scale optimizer training on a huge problem suite. It then evaluated the resulting optimizer on the VeLOdrome task suite  \cite{metz2022VeLO} and an early version of the MLCommons optimizer benchmark suite \cite{dahl2023benchmarking} where it claimed to provide decisive efficiency improvements over competitors, thus justifying its huge up-front training cost. This paper critically evaluates these claims.

%The complication for fairly benchmarking learned optimizers is accounting for the cost of meta-training stage. Usually, authors make amortization claims about meta-training cost. Nevertheless, any valid claim about amortization cost of meta-training must factor in the shift between meta-training and meta-testing distributions which in itself is ill-posed. For the sake of argument, say meta-training and meta-test distributions are identical, then meta-training cost can be directly compared to HPO results. Nevertheless, once a shift exists, the claim becomes a question of return on investment and direct comparisons with HPO become ill-defined without further assumptions.

\section{VeLO: Versatile Learned Optimizer}\label{sec:VeLO}
\keypoint{VeLO Architecture} VeLO \cite{metz2022VeLO} is a learned optimizer trained with the outer-objective (Eq~\ref{eq:outer}) of minimizing the training loss. The learned optimizer is parameterized as a hierarchical hypernetwork; a per-tensor LSTM that generates the parameters for a per-parameter MLP. The per-tensor hypernetwork operates on features aggregated from each parameter tensor, i.e: neural network layer. VeLO optimizer states and inputs include current iteration number, momentum at different timescales, squared gradients, adafactor-style accumulator, loss exponentially-moving average features, and tensors rank. 

\keypoint{VeLO Training} The meta-training task distribution included MLPs, CNNs, ResNets, ViTs, auto-encoders, variational auto-encoders, RNNs, and vanilla Transformers of various sizes. The architectures included dynamic configurations such as initialization and activation functions. Standard training datasets for image and language domains such as $16\times16$ ImageNet, CIFAR 10 and 100,  Fashion MNIST, LM1B, and Wikipedia English among others. The meta-optimizer used was standard evolutionary strategy  with antithetic-samples \cite{salimans2017evolution}. Meta-training spanned a total of 4000 TPU months with an online HPO procedure divided across 4 phases. Problem sizes and training unroll lengths were gradually increased over a curriculum which was found to improve meta-generalization.

\keypoint{VeLO Claims} Some key VeLO claims are (a) achieving a $4\times$ speedup over learning rate-tuned Adam on 50\% of tasks while being $16\times$ times faster on 14\% of VeLOdrome suite of tasks (\cite{metz2022VeLO}, Fig. 1). (b) out-performing hyperparameter tuned Adam on a suite of tasks from the \hyperlink{https://MLCommons.org/en/groups/research-algorithms/}{MLCommons algorithms track} in terms of the training loss (\cite{metz2022VeLO}, Sec. 4.2), (c) out-performing hyperparameter tuned Adam's generalization (validation loss) on the same benchmark (\cite{metz2022VeLO}, App. G.7). 

%The last claim is made by measuring the number of steps it takes learning rate tuned Adam to achieve the same loss VeLO reaches after 10K training steps.

It can be seen that VeLO's claims span learned optimizer benchmarking practical objectives of (a) training speedups and (b) absolute performance gains on both train and validation metrics while (c) meta-generalizing to new tasks distributions including VeLOdrome and MLCommons benchmark, which is a key justification behind amoritizing VeLO's meta-training cost.

\keypoint{Caveats} Besides the inputs discussed in the architecture paragraph above, VeLO needs one special input: It must be prompted with the total training steps it is expected to run for in order to initialize its states. This is then used to estimate  the fraction of training remaining online during learning.
%\doublecheck{In addition, VeLO requires total steps it will run  when intializing the optimizer states. This is used to compute the fraction of training remaining required as input to VeLO. 
For an explanation of how to control for this factor of variation fairly, we refer the reader to appendix \ref{app:VeLO_steps}.

\section{Benchmark Design}\label{sec:exp}
To examine VeLO's claims, our point of departure is the most recent time-to-result benchmark by MLCommons \cite{dahl2023benchmarking}. 
Comparing training curves to measure speedups is ill-posed. Therefore, the  MLCommons \cite{dahl2023benchmarking} {protocol measures learning speed by fixing a performance target (e.g., loss), and measuring time/steps taken for an optimizer to reach this target. 
Since VeLO also reported improved solution quality, we extend this protocol to the complementary perspective of fixing an optimization time/step budget %Not to be confused with MLCommons term budget.
and measuring the loss achieved at this point.}% (denoted performance control).

\keypoint{Baselines and Workloads} The original VeLO paper mainly compared with Adam. For more thorough evaluation, we train several GD variants, namely SGD with Heavy Ball Momentum, SGD with Nesterov Momentum, Adam, NAdam (Adam with Nesterov Momentum) and NAdamW. We train all baselines with default hyperparameters as reported in appendix \ref{app:def_hp}.
All algorithms are trained for a maximum allowed budget, either runtime or steps, on 4 workloads from the MLCommons benchmark, namely ResNet-50 on Imagenet, GNN on OGBG, DLRM on Criteo-1TB and U-Net on FastMRI. A workload is a fixed architecture, dataset and training objective. Please refer to \citep{dahl2023benchmarking} for details. %\doublecheck{Please note, optimizer capacity to overfit, train-val performance gap and training speedups are coupled. Since VeLO was trained to minimize the training loss rather than generalizing on an out-of-sample data, baselines are allowed to overfit for a fair comparison on both training and validation metrics.}

\keypoint{Measuring Training Speedups} The key evaluation hyperparameter in MLCommons is the notion of a \emph{performance target} (e.g., in units of loss) that defines a successful optimization. We can then measure speedups in terms of the wall-clock time or number of iterations taken to reach the target. 

Establishing performance targets is somewhat involved in the MLCommons methodology. First one sets a 
maximum allowed runtime in wall-clock or step count 
%budget (a maximum allowed runtime in wall-clock or step count)
for each workload, runs multiple trials of all algorithms for the full budget, and then measures the performance of all algorithms trials at 75\% of the maximum allowed budget. Then, for each algorithm on each workload, the median performance is selected, and the best performing algorithm defines the target for the workload. This translates to $\text{target}_w=\text{max}_{a}\{\text{median}_{s}\{L_{a, s, w}\}\}$ for all $w\in\mathcal{W}$ where $L_{a, s, w}$ is the performance metric of interest achieved by algorithm $a$ on trial $s$ and workload $w$. %and $s$ are all trials for a given algorithm. %Budget can be fixed wall-clock time or step count. We evaluate different targets for each protocol.

Subsequently we can measure the time/steps $t_{a,s,w}$ that the $s$th trial of any given algorithm $a$ takes to reach the target performance level $\textbf{target}_w$ on workload $w$. To aggregate results, we can employ performance profiles \cite{Dolan2002,dahl2023benchmarking}. Denote algorithms by $\mathcal{A}=\{a_1,a_2,..,a_k\}$ and workloads as $\mathcal{W}=\{w_1,w_2,..,w_n\}$. Then, given a workload $w$, we record the median time/steps taken for algorithm $a$ to achieve the performance target across all trials/seeds as $t_{a,w}=\textbf{median}_s\{t_{a,s,w}\}$. Then, to score an algorithm $\hat{a}$ on a workload $w$, the performance ratio is defined as:
\begin{equation}
    r_{\hat{a},w}=\frac{t_{\hat{a},w}}{\min_{a\in\mathcal{A}}t_{a,w}}
\end{equation}
The performance profile $\rho_{\hat{a}}(\tau)$ for an algorithm $\hat{a}$ on a random workload $w$ drawn uniformly from $\mathcal{W}$ is the probability of having a performance ratio $r_{\hat{a},w}$ of at most $\tau$:
\begin{equation}\label{eq:perf_score}
    \rho_{\hat{a}}(\tau)=\left(\frac{1}{n}\right) \times |\{w:r_{\hat{a},w}\leq\tau\}|
\end{equation}
Following \cite{dahl2023benchmarking}, the final score $B_a$ for each algorithm integrates the performance profile over a pre-defined range $r_{\textbf{max}}$ resembling a space of $\tau$s and normalized by $r_{\textbf{max}}-1$. This means that an algorithm that is consistently the fastest across all workload would have a score of 1.

\begin{wrapfigure}[12]{r}{0.43\columnwidth}
%\begin{figure}
%    \centering
    \includegraphics[width=1.0\linewidth]{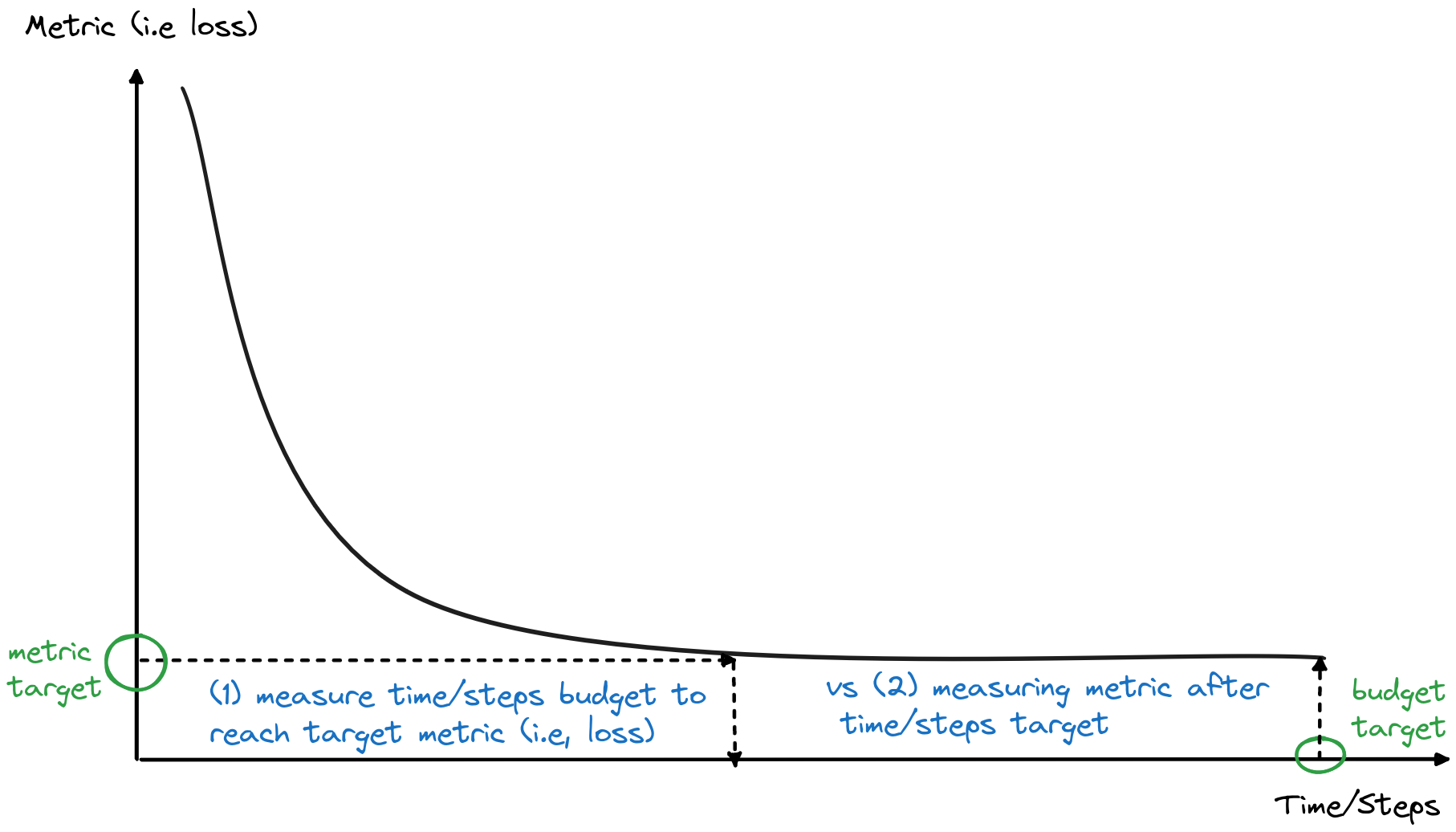}
    \caption{Illustration of optimizer learning metrics: Time/steps to performance target vs performance achieved at time budget.}
    \label{fig:schematic}
%\end{figure}
\end{wrapfigure}

In summary, for individual benchmarks $w$ and algorithm $a$, we report time-to-target $t_{a,w}$. We measure both wall-clock-time to target (denoted time-control condition), and steps to target (step-control condition). To aggregate across benchmarks we report the aggregate MLCommons score $B_a$. 

\keypoint{Measuring Training Quality} While MLCommons mainly focuses on training speedup, the complementary metric to is quality of solution found within a certain time/step budget. To this end, we also assess training and validation performance  $p_{a,w}$ (e.g., loss, accuracy) after algorithm $a$ reaching a certain time/step quota for workload $w$ (denoted performance-control).

For the specific workload budgets and targets found, please see Appendix~\ref{app:benchmark_details}. The training speedup vs training-quality metrics are illustrated schematically in Figure~\ref{fig:schematic}. Finally, see appendix \ref{app:eval_compar} for a comparison of similarities and extensions between our evaluation and VeLO's original assessment.

\section{Experiments}\label{sec:results}
%\paragraph{Position:} \textit{VeLO Scaling Cost Amortization Claim is Unjustified}

We now set out to assess whether VeLO's claims are justified, and hence whether its large up-front training cost can be justified from an amortization perspective.

Specifically, we ask the following questions:
\begin{description}
   \item [{(Q1)}] \emph{Is VeLO hyperparameter free as claimed?} 
  \item [{(Q2)}] \emph{Does VeLO indeed outperform existing hand-crafted optimizers on training and validation loss minimization as claimed?}
 \item [{(Q3)}] \emph{Does VeLO indeed provide dramatically faster optimization than standard baselines?}  
\end{description}

%We aim to provide corroboration evidence that challenges the justification for VeLO's scaling cost amortization claim. To summarize, we present findings on five key aspects. First, we show that VeLO exhibits hyperparameter sensitivity, with different runtime budget prompts influencing its acceleration behaviour. Second, despite claims of outperforming baselines in minimizing training loss, our results reveal that VeLO does not achieve this goal. Third,  we demonstrate that VeLO's acceleration in training targets is outperformed by baseline algorithms in both time and step budget experiments, indicating that VeLO does not consistently accelerate training. Fourth, VeLO's performance on validation loss does not significantly surpass that of untuned baselines. In light of these findings, we conclude that the meta-training of VeLO is unjustified on multiple levels, questioning the validity of its scaling cost amortization claim.

%TODO: summarize corroboration evidence (1) hyperparameter sensitivity, (2) outperformed on meta-learning goal of minimizing training loss, (3) not generalizing significantly better than baselines, and (4) having inconsistent speedups due to HP sensitivity. Add HPO results below to table \ref{tab:generalization}. Claim also that (5) VeLO and baselines are too far off from HPO results. Conclude that VeLO meta-training is unjustified on all levels.

%\paragraph{Conclusion 1:} \textit{VeLO is Not Hyperparameter Free but Hyperparameter Sensitive.}
\keypoint{Conclusion 1: VeLO is Not Hyperparameter Free but Hyperparameter Sensitive.}
Recall that the VeLO has one user-defined input: It requires prompting with the total number of steps (Sec~\ref{sec:VeLO} and \cite{metz2022VeLO}). It will accelerate (attempt to converge faster, but possibly reach a worse minima) if prompted with fewer steps. We study the MLCommons time-to-performance target protocol for different values of this hyperparameter. We consider prompting with steps corresponding to either 100\%  or 75\% of MLCommons wall-clock max runtime. From the results in Table~\ref{tab:budget_speed} we see that the prompt is actually a key hyperparameter. For example, the 75\% prompt reaches the Criteo training target before timeout, while the 100\% prompt doesn't succeed in time. Meanwhile the 100\% prompt reaches the OGBG training target before timeout, while the 75\% prompt does not (it behaves too greedily and converges to a poor optimum worse than the required performance target). Overall VeLO is in fact sensitive to the number of steps hyperparameter, often crucially so. 

%In a time-controlled setting, we prompted VeLO with as many steps as the wall-clock time budget allowed. We also conducted additional runs with a step limit corresponding to 75\% of the wall-clock time budget. Results presented in table \ref{tab:budget_speed} demonstrate that VeLO's  time to result differs across prompts. This suggests that VeLO uses the prompt as a key hyperparameter, influencing its acceleration behaviour. The shorter the budget, the faster VeLO accelerates validation targets. Meanwhile, different budgets are needed to reach different training targets. With furhter corroborating evidence, table \ref{tab:budget_val} shows that VeLO's performance remains consistent across budgets. With less compute, VeLO produces fewer gradient updates, effectively regularizing the training loss. This explains the slight improvement in generalization. Thus, tuning the budget prompt significantly affects VeLO's behavior.

\begin{table}[h]
    \centering
    \caption{Influence of VeLO's step prompt hyperparameter. Time ($\downarrow$, sec) taken to reach MLCommons training and validation performance targets. The prompt is given as steps corresponding to 100\% or 75\% of MLCommons wall-clock budget for the task. (-) indicates timeout. }%Prompting VeLO with a different runtime budgets influences it's acceleration to targets. %The \textit{shorter run} is 75\% of the runtime wall-clock budget used for \textit{baseline}. The shorter the run, the faster it reaches the target.}
    \label{tab:budget_speed}
    \begin{tabular}{cccccc}
         \toprule
        & Workload & Criteo & FastMRI & ImageNet & OGBG \\
%        Optimizer &  &  &  &  \\
        \midrule
%        & \multirow{2}{*}{\rotatebox{90}{Train}}   & \\
%        \midrule
        \multirow{2}{*}{\rotatebox{90}{Train}}  & VeLO (100\%) & - & - & - & \textbf{13215} \\
        & VeLO (75\%)  & \textbf{6088} & - & - & - \\
        \midrule
        %& \multicolumn{2}{c}{Validation Target}  & \\
 %       \midrule
        \multirow{2}{*}{\rotatebox{90}{Val}}& VeLO (100\%)  & - & \textbf{5728} & 62587 & 8779 \\
        &VeLO (75\%)  & - & 5913 & \textbf{58713} & \textbf{7238} \\
        \bottomrule
    \end{tabular}
\end{table}

% \begin{table}[h]
%     \centering
%     \caption{VeLO produces comparative performances across budgets on both training and validation losses. The \textit{shorter run} is 75\% of the runtime wall-clock budget used for \textit{baseline}. Shorter run better generalization is induced by the implicit regularization of smaller budget; similar to early stopping.}
%     \label{tab:budget_val}
%     \begin{tabular}{ccccc}
%          \toprule
%         Workload & Criteo & FastMRI & ImageNet & OGBG \\
%         Optimizer &  &  &  &  \\
%         \midrule
%          &  & Training Loss & & \\
%         \midrule
%         Baseline & 0.1232 $\pm$ 0.00039 & 0.2737 $\pm$ 0.00324 & 0.0862 $\pm$ 0.00715 & 0.0164 $\pm$ 0.00037 \\
%         Shorter Run & 0.1236 $\pm$ 0.00024 & 0.2763 $\pm$ 0.00128 & 0.1445 $\pm$ 0.00158 & 0.0180 $\pm$ 0.00058 \\
%         \bottomrule
%         &  & Validation Loss & & \\
%         \midrule
%         Baseline & 0.1240 $\pm$ 0.00005 & 0.2851 $\pm$ 0.00008 & 1.5358 $\pm$ 0.02996 & 0.0510 $\pm$ 0.00017 \\
%         Shorter Run & 0.1242 $\pm$ 0.00003 & 0.2850 $\pm$ 0.00026 & 1.3775 $\pm$ 0.00333 & 0.0491 $\pm$ 0.00064 \\
%         \bottomrule
%     \end{tabular}
% \end{table}

%\paragraph{Conclusion 2:} \textit{VeLO Does Not Outperform Baselines in Minimizing Both Training and Validation Losses.}
\keypoint{Conclusion 2: VeLO Does Not Outperform Baselines in Minimizing Both Training and Validation Losses.}
VeLO reported outperforming hand-crafted optimizers in terms of achieving lower training and validation losses, both on VeLOdrome and MLCommons (algorithm) benchmark suites. But the associated experiments on MLCommons compared against Adam alone \cite{metz2022VeLO}. Meanwhile, \cite{dahl2023benchmarking} observe that different optimizers often `win' on different benchmarks. So we directly compare VeLO against a range of off-the-shelf optimizers with default hyperparameters in terms of optimization quality after a fixed step-budget on MLCommons. From the results in Table~\ref{tab:generalization} (see full details in Appendix~\ref{app:step_controlled}), we see a different picture: VeLO is not a consistent winner in either train or validation loss achieved, despite that we conducted no HPO at all on the baselines. We attribute this discrepancy two factors: (1) VeLO \cite{metz2022VeLO} evaluating insufficient competitors in their original comparison -- since as we see different competitors win on different benchmarks/metrics. (2) VeLO's evaluation primarily focused on VeLOdrome benchmarks, which were reportedly more similar to VeLO's training distribution \cite{metz2022VeLO}, and focused less on the MLCommons suite which was reportedly more different. To the extent that this is the explanation, it suggests that VeLO is not as general purpose as claimed, and thus undermines the amortization argument used to justify its up-front training cost. 

%Optimizers allowed to overfit outperform VeLO's main meta-training goal. Moreover, although originally VeLO's claim was outperforming Adam HPO trials on validation performance, we find no evidence of this. VeLO performance is not significantly better than untuned baselines on the validation set.

\begin{table}[tb]
    \centering
    \caption{Training and validation losses $(\downarrow)$ across workloads after training each algorithm for a fixed number of steps. VeLO is not a consistent winner. }% Note comparative performances across all algorithms including VeLO.}
    \label{tab:generalization}
    \resizebox{\textwidth}{!}{
    \begin{tabular}{cccccc}
        \toprule
        & Workload & Criteo-1TB& FastMRI & ImageNet & OGBG \\
        & Optimizer &  &  &  &  \\
        \midrule
        \multirow{6}{*}{\rotatebox{90}{Train}} &
        Adam & 0.1225 $\pm$ 0.00015 & \textbf{0.2702 $\pm$ 0.00693} & 0.0470 $\pm$ 0.00587 & 0.0165 $\pm$ 0.00033 \\
        &Heavy Ball & 0.1293 $\pm$ 0.00045 & 0.2809 $\pm$ 0.00282 & 0.2582 $\pm$ 0.02502 & 0.0341 $\pm$ 0.00026 \\
        &NAdam & 0.1239 $\pm$ 0.00302 & 0.2692 $\pm$ 0.00396 & 0.0571 $\pm$ 0.00060 & 0.0173 $\pm$ 0.00033 \\
        &NAdamW & \textbf{0.1220 $\pm$ 0.00018} & 0.2750 $\pm$ 0.00102 & \textbf{0.0470 $\pm$ 0.00106} & 0.0197 $\pm$ 0.00196 \\
        &Nesterov & 0.1301 $\pm$ 0.00096 & 0.2811 $\pm$ 0.00505 & 0.2512 $\pm$ 0.02606 & 0.0331 $\pm$ 0.00008 \\
        &VeLO & 0.1229 $\pm$ 0.00034 & 0.2804 $\pm$ 0.00008 & 0.1046 $\pm$ 0.00342 & \textbf{0.0153 $\pm$ 0.00088} \\
        \midrule
         \multirow{6}{*}{\rotatebox{90}{Validation}} &
        Adam & \textbf{0.1237 $ \pm $ 0.00005} & \textbf{0.2850 $ \pm $ 0.00003} & 1.9438 $ \pm $ 0.00717 & 0.0515 $ \pm $ 0.00020 \\
        &Heavy Ball & 0.1299 $ \pm $ 0.00062 & 0.2899 $ \pm $ 0.00008 & 1.6870 $ \pm $ 0.02763 & 0.0466 $ \pm $ 0.00038 \\
        &NAdam & 0.1256 $ \pm $ 0.00315 & 0.2851 $ \pm $ 0.00022 & 1.9528 $ \pm $ 0.00120 & 0.0509 $ \pm $ 0.00027 \\
        &NAdamW & \textbf{0.1237 $ \pm $ 0.00005} & 0.2851 $ \pm $ 0.00016 & 1.6345 $ \pm $ 0.01098 & 0.0483 $ \pm $ 0.00157 \\
        &Nesterov & 0.1305 $ \pm $ 0.00067 & 0.2899 $ \pm $ 0.00004 & 1.7022 $ \pm $ 0.03822 & \textbf{0.0462 $ \pm $ 0.00027} \\
        &VeLO & 0.1240 $ \pm $ 0.00008 & 0.2851 $ \pm $ 0.00022 & \textbf{1.5017 $ \pm $ 0.02693} & 0.0522 $ \pm $ 0.00126 \\
        \bottomrule
        \end{tabular}}
\end{table}

%\paragraph{Conclusion 3:} \textit{VeLO Does Not Provide Faster Training.}
\keypoint{Conclusion 3: VeLO Does Not Provide Faster Training.}
VeLO claims substantially faster training. It was trained for the objective of fast training loss minimisation, and empirically observed to also provide fast validation loss minimisation. However, again these original claims were largely based on the  VeLOdrome benchmark (which may be unrealistically easy, as discussed in the previous section), and in terms of MLCommons they were based on comparison to Adam alone. We now compare VeLO to a range of off-the-shelf optimizers with default hyperparameters on our four MLCommons tasks using the time/steps to performance target protocol of MLCommons. The MLCommons benchmark results 
%Hyperparameter dependence of VeLO influences observed speedups. Scaling learned optimizer training taught VeLO various learning rate schedules. Prompting VeLO with different budgets probes this knowledge. To examine VeLO's speedup claims under both time-to-result and step-to-result metrics while fixing wall-clock time and total step budgets, respectively. Scores 
presented in Table~\ref{tab:performance_scores} in terms of the aggregate MLCommons score $B_a$, which integrates over the algorithms' performance profiles (see Appendix \ref{app:time_controlled} and \ref{app:step_controlled} for details).

Surprisingly, VeLO is far from best in training speed (which might be expected given it is optimised for training efficacy), although it surpasses some baselines in speed of minimising the validation loss. VeLO's loss to Adam in training efficiency we attribute to (1) weak generalisation to the MLCommons task suite, and (2) Adam's default learning rate decay schedule potentially being more effective than the outcome of the amount of HPO applied with Adamin in \cite{metz2022VeLO}. VeLO's comparative success in validation is potentially attributable to several MLcommons workloads being in the overfitting regime\footnote{A regime where fully minimizing the training loss ultimately worsens validation performance}, so  VeLO's less effective minimisation of the train loss can lead to better validation than competitors. (Note that while we measure validation performance, all optimizers are run with default parameters and not tuned on validation metrics.). This is particularly the case in the time-control condition because since Velo is slower per-iteration than the baselines, it runs fewer iterations than baselines when using a wall clock-time budget, and thus effectively benefits from early stopping compared to the baselines. Finally, returning to the hyperparameter sensitivity issue from Experiment 1, we also compare VeLO with 75\% of the total step-prompt and see a noticeable impact in the score distribution. 
\textit{}\begin{wraptable}[16]{r}{0.52\columnwidth}
%\begin{table}[h]
    \centering
    \caption{Optimizer speed evaluation (MLCommons score $B_a, (\uparrow)$, Eq.~\ref{eq:perf_score}). Time-To-Result and Steps-To-Results are reported when fixing wall-clock time and steps respectively across train and validation targets.\\}
    \label{tab:performance_scores}
    \begin{tabular}{cccccc}
    \toprule
        \textbf{Optimizer} & \multicolumn{2}{c}{\textbf{Train Scores}} & \multicolumn{2}{c}{\textbf{Validation Scores}} \\
        \cmidrule(lr){2-3} \cmidrule(lr){4-5}
         & \textbf{Time} & \textbf{Step} & \textbf{Time} & \textbf{Step} \\
        \midrule
        NAdam & 0.24 & 0.25 & 0.00 & 0.23 \\
        NAdamW & 0.36 & 0.25 & 0.49 & 0.36 \\
        Adam & \textbf{1.00} & \textbf{1.00} & 0.24 & \textbf{0.50} \\
        Nesterov & 0.00 & 0.00 & 0.00 & 0.00 \\
        HeavyBall & 0.00 & 0.00 & 0.00 & 0.00 \\
        VeLO & 0.19 & 0.00 & \textbf{0.71} & 0.39 \\
        \midrule
        VeLO Short & 0.16 & 0.03 & \underline{0.74} & \underline{0.59} \\
        \bottomrule
    \end{tabular}
%\end{table}
\end{wraptable}

\section{Conclusion} Learned optimizers have shown substantial success on narrowly defined task distributions. VeLO scaled up optimizer learning to train a foundational optimizer on a vast task distribution at huge cost. The vision was that it would then generalize to arbitrary machine learning workloads, and outperform hand-crafted optimizers, thus justifying its up-front training cost. We were initially optimistic and excited to see this in action. However ultimately, our independent evaluation on the MLCommons optimizer benchmark called into question most of VeLO's big claims of being hyperparameter free, and providing improved and faster optimization. 

\newpage
\begin{ack}
We extend sincere gratitude to Frank Schneider, Zach Nado, and George Dahl for their support. During the course of this research, they provided clarifications on conceptual ideas regarding the design and implementation of the MLCommons benchmark.

The project was supported by the Royal Academy of Engineering under the Research Fellowship programme. This work was also supported by the Edinburgh International Data Facility (EIDF) and the Data-Driven Innovation Programme at the University of Edinburgh. 
\end{ack}

\bibliography{references}
\newpage
\appendix
\section{Benchmark Details}\label{app:benchmark_details}
\keypoint{Setting Maximum Allowed Wall-Clock Time Across Hardware} The MLCommons benchmark is based on fixing a maximum allowed wall-clock time for each workload, denoted time-control condition. To transfer this maximum allowed wall-clock runtime across hardware, we compute the ratio between time per step of algorithms on their hardware (8$\times$V100) and ours (1$\times$A100-80GB or 2$\times$A100-80GB). Then, this ratio is used as a multiplier factor of the maximum allowed wall-clock time for each workload.

To get time per step for the original V100 hardware, we use total number of steps each algorithm runs for and the equivalent wall-clock time for those steps as supplemented by the authors in table 28 in \cite{dahl2023benchmarking}. For a given workload, the bold time entry is the maximum allowed wall-clock runtime. For the algorithm with this bold-entry, the steps it runs for in the wall-clock time can be found in the \textit{Steps} row. For reference, we copy the numbers here as in table \ref{tab:8v100_runtimes}.

\begin{table}[h]
    \centering
    \caption{Steps and corresponding runtime for the target-setting algorithms from \cite{dahl2023benchmarking} table 28 on (8 x V100 16GB) Hardware.}
    \label{tab:8v100_runtimes}
    \begin{tabular}{lcccc}
         \toprule
         \multicolumn{5}{c}{Workload Datasheet for (8 x V100 16GB) Hardware}\\
         \midrule
         &\textbf{Criteo 1TB}	& \textbf{FastMRI} &	\textbf{ImageNet} &	\textbf{OGBG}\\
         & \textbf{DLRMsmall} &	\textbf{U-Net} &	\textbf{Resnet-50} &	\textbf{GNN} \\
         \midrule
         \textbf{Optimizer} &	NAdamW &	Nesterov &	Heavy Ball &	Nesterov\\
         \textbf{Steps} &	10,667 &	38,189 &	186,667 &	80,000 \\
         \textbf{Runtime (sec}) &	7703	& 8859 &	63,008 &	18,477 \\
         \textbf{Time Per Step (sec)} &	0.7221336833 &	0.2319777947	& 0.3375422544 &	0.2309625\\
         \bottomrule
    \end{tabular}
\end{table}

To transfer the wall-clock time, we execute the implementation of the algorithm with this bold time entry as found  \hyperlink{https://github.com/MLCommons/algorithmic-efficiency/tree/main/reference_algorithms/target_setting_algorithms}{here} on our hardware for 5\% of it's steps. Then, we compute our time-per-step on both $1\times$ and $2\times$ A100 GPUs. Finally, we use the ratio between the V100 and A100 time-per-step as a factor multiplication of the wall-clock time. The time per step hardware benchmarking results for A100 GPUs are shown in table \ref{tab:runtimes_a100}. To maximally utilize our infrastructure, we use 2 GPUs for FastMRI and ImageNet workloads and 1 GPU for Criteo and OGBG experiments. The final maximum allowed wall-clock runtime are also reported in table \ref{tab:runtimes_a100}.

\begin{table}[h]
    \centering
    \caption{Steps and corresponding runtime for the target-setting algorithms on A100-SXM4-80GB Hardware. The steps FastMRI workload is benchmarked for are 1101 steps less than the 5\% ratio. This is because FastMRI data loading introduces initial noise in the algorithm and hence time per step are evaluated over the final 808 steps.}
    \label{tab:runtimes_a100}
    \begin{tabular}{p{4cm}cccc}
        \toprule
        &\textbf{Criteo 1TB}	& \textbf{FastMRI} &	\textbf{ImageNet} &	\textbf{OGBG}\\
        & \textbf{DLRMsmall} &	\textbf{U-Net} &	\textbf{Resnet-50} &	\textbf{GNN} \\
        \midrule
        \textbf{Steps ($\sim$5\% of table \ref{tab:8v100_runtimes})} & 533 & 808 & 9,000 & 4,000\\
         \midrule
         \multicolumn{5}{c}{$1\times$A100-SXM4-80GB Hardware}\\
        \midrule
        \textbf{Runtime (sec)} & 569 & 367 & 8,387 & 743 \\
        \textbf{Time Per Step (sec)} & 1.067542214 & 0.4542079208 & 0.9318888889 & 0.185745\\
        \midrule
        \multicolumn{5}{c}{$2\times$A100-SXM4-80GB Hardware}\\
        \midrule
        \textbf{Runtime (sec)} & 513 & 220 & 3,951 & 666\\
        \textbf{Time Per Step (sec)} & 0.9624765478 & 0.2699386503 & 0.4381237525 & 0.1665\\
        \midrule
        \multicolumn{5}{c}{Final Maximum Allowed Wall-Clock Times}\\
        \midrule
        \textbf{Number of GPUs} & 1 & 2 & 2 & 1\\
        \textbf{Maximum Allowed Time} & 11,387 & 10,308 & 81,783 & 14,859\\
        \bottomrule
    \end{tabular}
\end{table}

\keypoint{Established Targets in Maximum Allowed Time/Steps} We set targets and measure time/steps to reach those targets as standardized in the MLCommons benchmark. To set the self-tuning regime targets, we use the methodology introduced in section \ref{sec:exp} as done originally by \cite{dahl2023benchmarking}. We set separate targets for the time-control and step-control conditions. Targets and maximum allowed runtimes for both time-control and step-control conditions in table \ref{tab:targets}. The tables also include the maximum allowed wall-clock time or maximum allowed steps to run for.

\begin{table}[h]
    \centering
    \caption{Total allowed runtimes and targets for all workloads. The target metrics are cross-entropy loss (CE), structural similarity index measure (SSIM), accuracy and mean average precision (mAP)}
    \label{tab:targets}
    \begin{tabular}{lcccc}
        \toprule
        &\textbf{Criteo 1TB}	& \textbf{FastMRI} &	\textbf{ImageNet} &	\textbf{OGBG}\\
        & \textbf{DLRMsmall} &	\textbf{U-Net} &	\textbf{Resnet-50} &	\textbf{GNN} \\
        \midrule
        \multicolumn{5}{c}{Wall-Clock Time-Control Condition}\\
        \midrule
        Maximum Allowed Time & 11,387 & 10,308 & 81,783 & 14,859\\
        Metric & CE & SSIM & Accuracy & mAP\\
        Training Target & 0.12215 & 74.543\% & 98.709\% & 75.946\% \\
        Validation Target & 0.12367 & 72.671\% & 71.012\% & 27.867\% \\
        \midrule
        \multicolumn{5}{c}{Step-Control Condition}\\
        \midrule
        Maximum Allowed Steps & 7,000 & 36,000 & 150,000 & 80,000\\
        Metric & CE & SSIM & Accuracy & mAP\\
        Training Target & 0.1225 & 74.535\% & 98.147\%& 76.547\%\\
        Validation Target & 0.12408 & 72.659\% & 70.600\%& 27.849\%\\
        \bottomrule
    \end{tabular}
\end{table}

\keypoint{Measuring Training Quality} For measuring training quality, we measure the final performance $p_{a,w}$ achieved within a certain time/step budget. The time and steps budgets used are the maximum allowed wall-clock time and maximum allowed steps used for the time-to-result benchmark. These maximum runtime are presented in table \ref{tab:targets} as maximum allowed steps and maximum allowed time for step-control and time-control conditions respectively.

\section{Managing VeLO Inputs}\label{app:VeLO_steps}
VeLO requires total steps at input to intialize optimizer states. This is used to compute percentage of remaining training, a feature input to the LSTM hypernetwork. We can follow two different approaches to provide this input to VeLO. First, we could refactor the benchmark and VeLO implementation to provide the percentage of time remaining directly as input while the experiment is running. This would be computed as the ratio of time remaining and total allowed runtime. The remaining time can be computed directly from the benchmark \href{https://github.com/MLCommons/algorithmic-efficiency/blob/main/submission_runner.py}{accumulated\_submission\_time} variable from the MLCommons benchmark which is updated every step by the \href{https://github.com/MLCommons/algorithmic-efficiency/blob/main/algorithmic_efficiency/profiler.py}{profiler}. A simpler approach is estimating the steps VeLO can take within the maximum allowed wall-clock time. We opt for the latter. In table \ref{tab:VeLO_input}, we provide the estimates over two runs of VeLO for 5\% of the step hints discussed in appendix \ref{app:benchmark_details}.

The implementation is written in jax. Jax compiles the computational graphs using XLA. We ommit the compilation times from the estimates since they take insignificant ratio from the whole training runtime but can potentially influence the estimate over 5\% of the runtime. To explain the rows in table \ref{tab:VeLO_input}, we first run VeLO for a fixed number of steps corresponding to row \textbf{Steps Run}. These are the same steps used earlier in appendix \ref{app:benchmark_details}. Then, we measure the total runtime as reported in \textbf{Observed Runtime (sec)}. The \textbf{Time Per Step (sec)}, the ratio of the first and second row are used to estimate the hyperparameter, \textbf{Estimated Total Steps}. Subsequently, we average the total steps VeLO can fit in the runtime over the two estimates. We train the workloads using VeLO from start to finish once and then update the total steps for each workload given the actual observed steps and run for two more trials. Since we take median over trials, the evaluation of VeLO is insensitive to any outliers produced by the estimates.

\begin{table}[h]
    \centering
    \caption{VeLO main hyperparameter estimate; total steps it will run for.}
    \label{tab:VeLO_input}
    \begin{tabular}{p{4cm}cccc}
         \toprule
        &\textbf{Criteo 1TB}	& \textbf{FastMRI} &	\textbf{ImageNet} &	\textbf{OGBG}\\
        & \textbf{DLRMsmall} &	\textbf{U-Net} &	\textbf{Resnet-50} &	\textbf{GNN} \\
        \midrule
        \textbf{Number GPUs}	 &1&	2	&2&	1\\
        \midrule
        \multicolumn{5}{c}{VeLO Total Steps Estimate 1}\\
        \midrule
        \textbf{Steps Run}	&533&	44,799&	9,018&	4,000\\
        \textbf{Observed Runtime (sec)}	& 791	&12080&	4651	&750\\
        \textbf{Time Per Step (sec)}	&1.484052533&	0.2696488761&	0.5157462852&	0.1875\\
        \textbf{Estimated Total Steps} &	7672	& 38227 &	158572 &	79248\\
        \midrule
        \multicolumn{5}{c}{VeLO Total Steps Estimate 2}\\
        \midrule
        \textbf{Steps Run} &	533 &	65,576 &	9,018 &	4,000\\
        \textbf{Observed Runtime (sec)} &	791	& 17863 &	4635 &	746\\
        \textbf{Time Per Step (sec)} & 1.484052533	&0.2724014883	&0.5139720559	&0.1865\\
        \textbf{Estimated Total Steps}	& 7672	&37841	&159119	&79672\\
        \midrule
        \textbf{Trial 1 Estimated Steps} &	7672 &	38034 &	158845 &	79460\\
        \textbf{Trial 1 Actual Steps (used in trials 2/3)} &7545&37160&156960&74972\\
        \midrule
    \end{tabular}
\end{table}
For the performance-control condition, where we run for a total fixed number of steps, we run VeLO from start to finish given the maximum allowed steps in table \ref{tab:targets}. Meanwhile, for the VeLO Short run, denoted also as VeLO (75\%) in table \ref{tab:budget_speed}, which is prompted with 75\% with the steps VeLO can run in the maximum allowed wall-clock time, we use 75\% of the steps reported in the final row of table \ref{tab:VeLO_input}.

\section{Default Hyperparameters}\label{app:def_hp}
\begin{figure}[h]
    \centering
    \includegraphics[width=\linewidth]{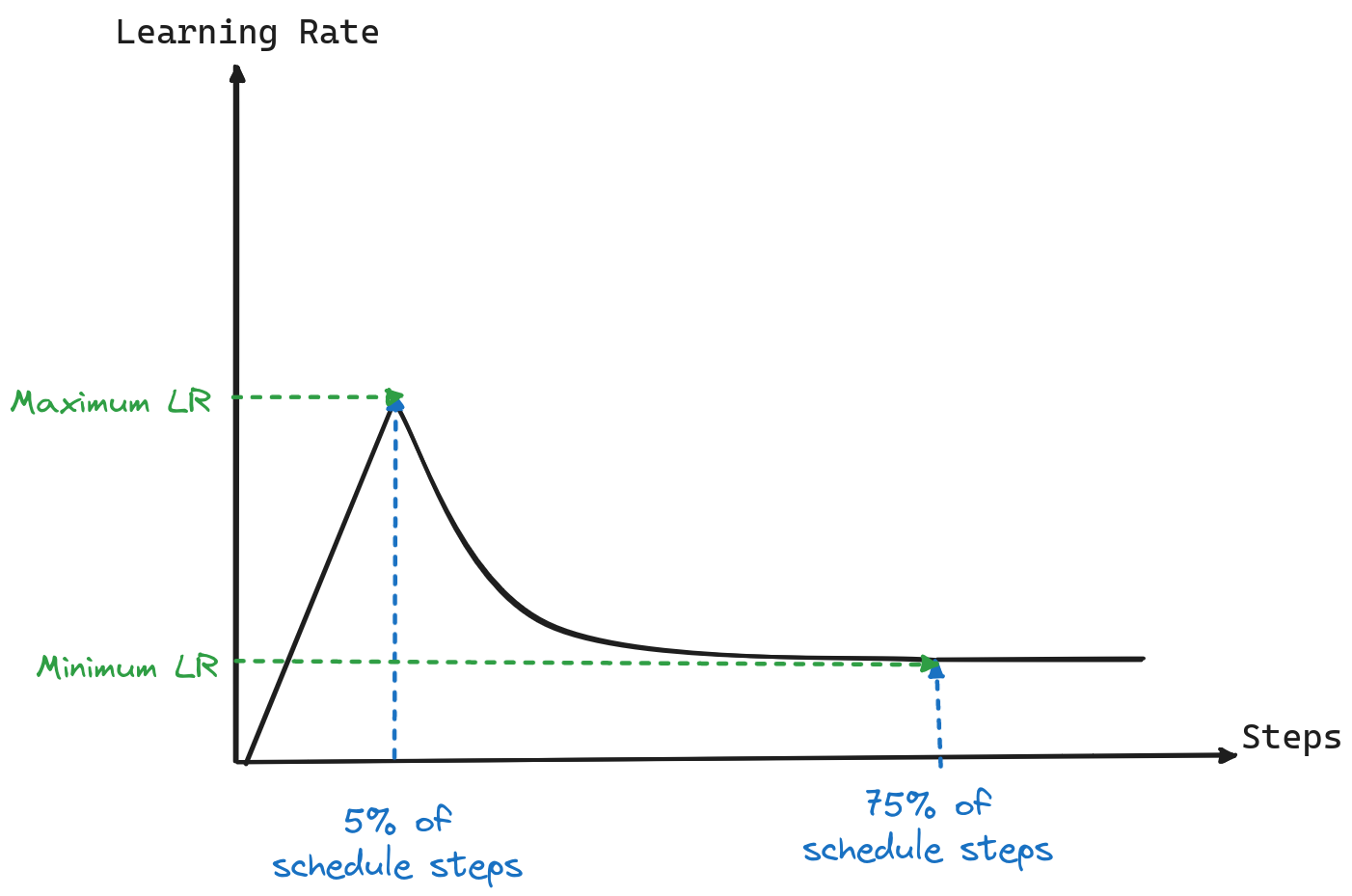}
    \caption{The Learning Rate schedule used for Adam variants from table \ref{tab:def_hp}. The first 5\% of steps are the linear warmup to the maximum LR, which is followed by 75\% of cosine decay to a minimum LR.}
    \label{fig:lr_sch}
\end{figure}

For all default hyperparameters used for Adam and SGD variants, please refer to table \ref{tab:def_hp}. The learning rate schedule consists of a linear warmup followed by cosine decay as illustrated in figure \ref{fig:lr_sch}. The schedule requires a total number of steps to operate on. We set the total steps of the schedule to 75\% of the step hint provided by the MLCommons benchmark for each workload. The step hint is approximately the total steps the SGD variants can run for given the maximum allowed wall-clock time of the benchmark. We set the warmup and cosine decay steps to the first 5\% of the schedule steps and the remaining 95\% respectively.
\begin{table}[h]
    \centering
     \caption{Default Hyperparameters for all baseline optimizers.}
    \label{tab:def_hp}
    \begin{tabular}{cccccc}
        \toprule
         Hyperparameter & Adam & NAdamW & NAdam & Heavy Ball & Nesterov   \\
         \midrule
         Base LR & 5e-3 & 1e-2 & 1e-2 & 5e-2 & 5e-2 \\
         $L_2$ Regularization & - & 4e-3 & - & - & - \\
         $\beta_1$ & 0.9 & 0.9 & 0.9 & 0.9 & 0.9 \\
         $\beta_2$ & 0.999 & 0.999 & 0.999 & - & - \\
         Schedule & linear warmup & linear warmup & linear warmup & - & -  \\
         & + cosine decay & + cosine decay & + cosine decay & - & - \\
         Warmup Steps & 5\% & 5\% & 5\% & - & - \\
         Decay Steps & 95\% & 95\% & 95\% & - & - \\
         Minimum LR & 1e-4 & 1e-4 & 1e-4 & - & - \\
         \bottomrule
    \end{tabular}
\end{table}

\clearpage
\section{Time-Controlled Experiments Results}\label{app:time_controlled}
\subsection{Time-To-Result Measurements}
\begin{table}[h]
    \centering
    \caption{Total wall-clock time (sec) to achieve the train target performance. These numbers are used to plot the performance profiles in figure \ref{fig:time_train_profile}. A value of inf indicates that that baseline was unable to achieve the target within the maximum allowed runtime.}
    \begin{tabular}{lcccc}
        \toprule
        Workload & Criteo & FastMRI & ImageNet & OGBG \\
        Optimizer &  &  &  &  \\
        \midrule
        Adam & \textbf{3746} & \textbf{5817} & \textbf{48341} & \textbf{9277} \\
        HeavyBall & inf & inf & inf & inf \\
        NAdam & 4096 & inf & inf & inf \\
        NAdamW & 6232 & inf & 70831 & inf \\
        Nesterov & inf & inf & inf & inf \\
        VeLO & inf & inf & inf & 13215 \\
        \midrule
        VeLO Short & 6088 & inf & inf & inf \\
        \bottomrule
    \end{tabular}
\end{table}

\begin{figure}[h]
    \centering
    \caption{Time-to-Target performance profiles of baselines vs VeLO and VeLO with 75\% (VeLO Short) prompt on training targets.}
    \label{fig:time_train_profile}
    \includegraphics[width=\linewidth]{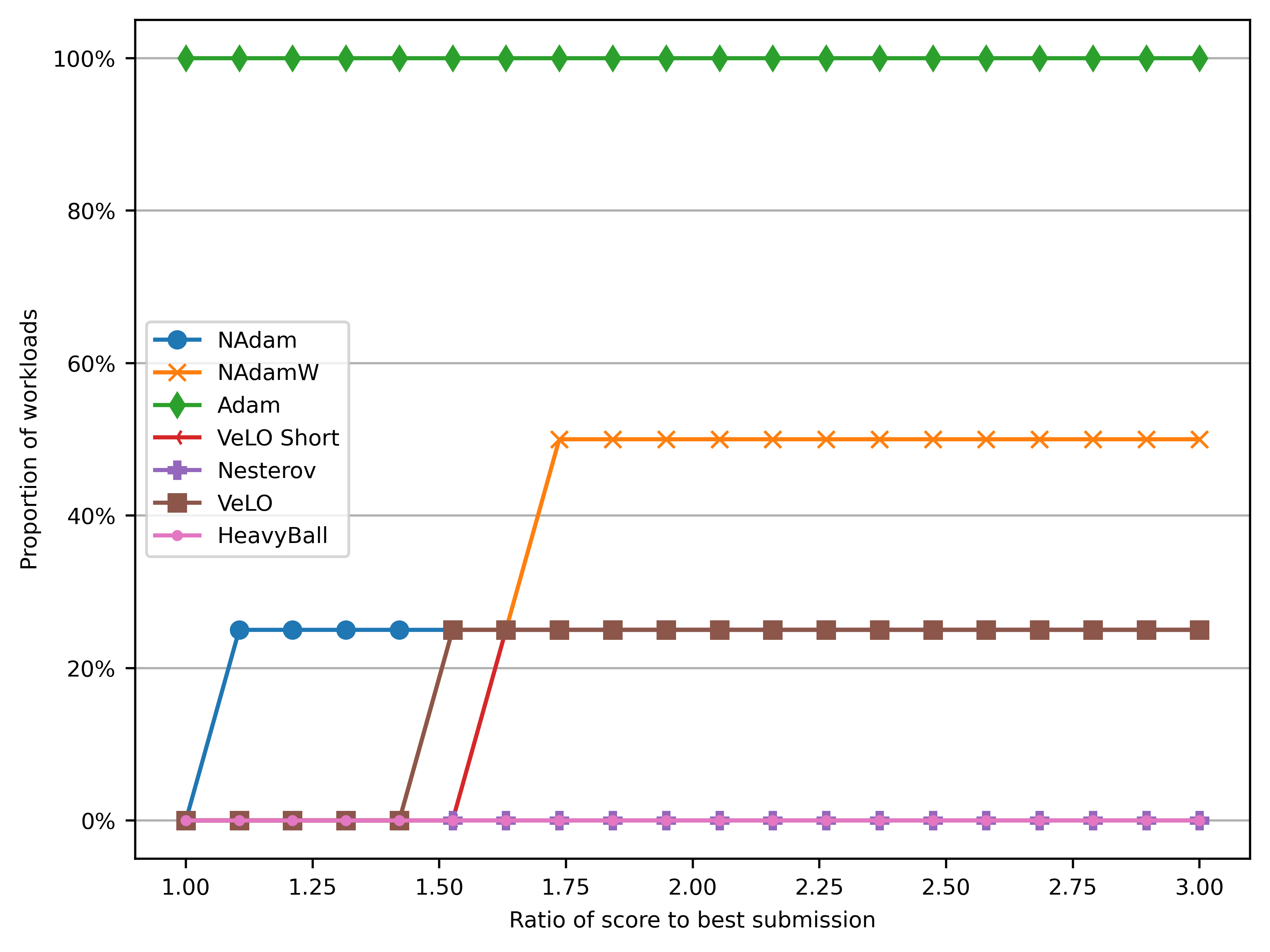}
\end{figure}

\begin{table}[h]
    \centering
    \caption{Total wall-clock time (sec) to achieve the validation target performance. These numbers are used to plot the performance profiles in figure \ref{fig:time_val_profile}. A value of inf indicates that that baseline was unable to achieve the target within the maximum allowed runtime.}
    \begin{tabular}{lcccc}
        \toprule
        Workload & Criteo & FastMRI & ImageNet & OGBG \\
        Optimizer &  &  &  &  \\
        \midrule
        Adam & 7644 & inf & inf & inf \\
        HeavyBall & inf & inf & inf & inf \\
        NAdam & inf & inf & inf & inf \\
        NAdamW & \textbf{6939} & inf & 63502 & inf \\
        Nesterov & inf & inf & inf & inf \\
        VeLO & inf & \textbf{5728} & \textbf{62587} & \textbf{8779} \\
        \midrule
        VeLO Short & inf & 5913 & \underline{58713} & \underline{7238} \\
        \bottomrule
    \end{tabular}
\end{table}

\begin{figure}[h]
    \centering
    \caption{Time-to-Target performance profiles of baselines vs VeLO and VeLO with 75\% (VeLO Short) prompt on validation targets.}
    \label{fig:time_val_profile}
    \includegraphics[width=\linewidth]{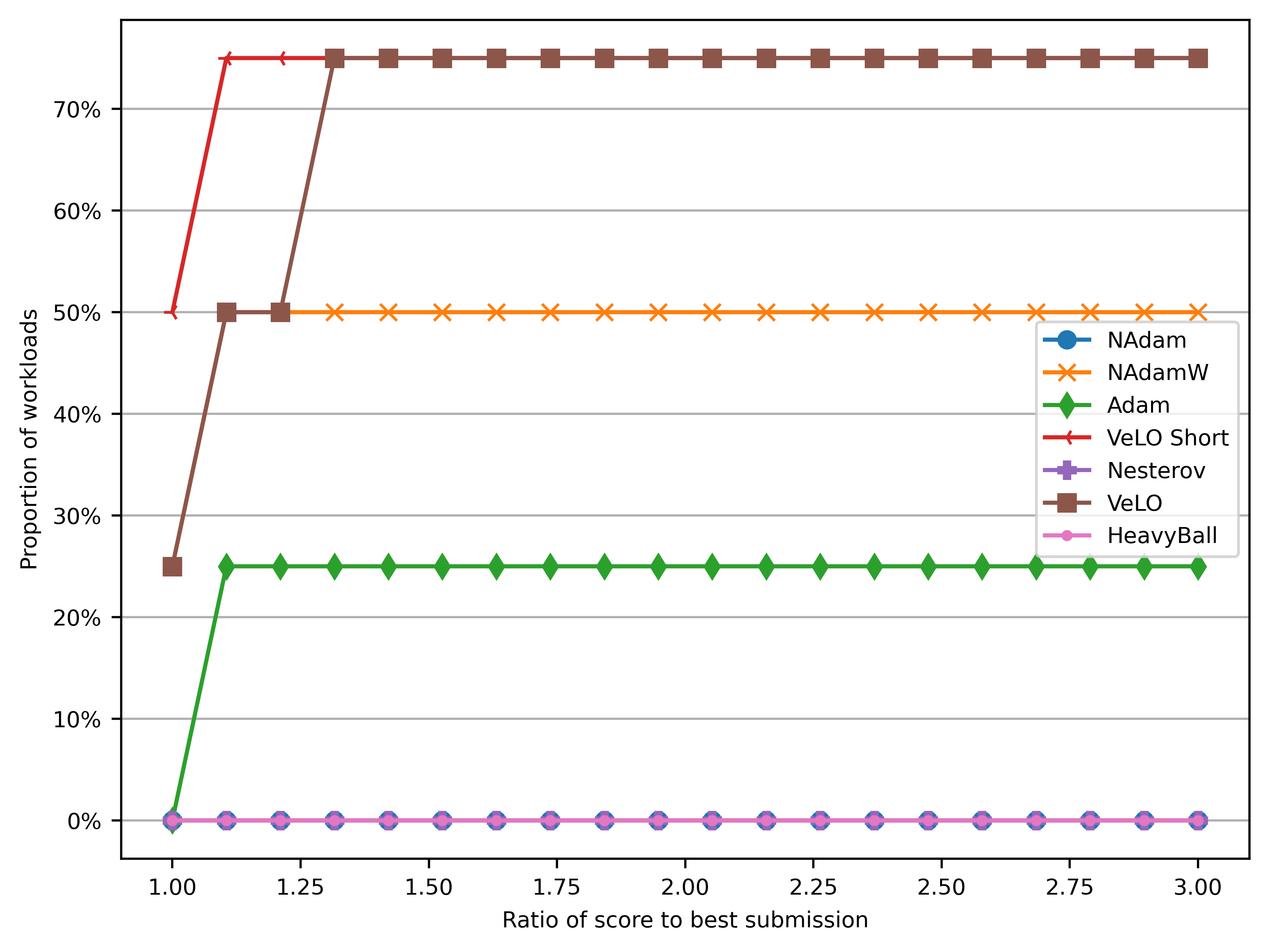}
\end{figure}

\subsection{ImageNet}
\begingroup
\begin{tabular}{lcccc}
\hline
Name & train/loss & train/accuracy & validation/loss & validation/accuracy \\
\hline
Adam & 0.0481 $\pm$ 0.00531 & 98.6401 $\pm$ 0.16370 & 1.9405 $\pm$ 0.00915 & 69.8780 $\pm$ 0.03027 \\
Heavy Ball & 0.1961 $\pm$ 0.06906 & 94.1552 $\pm$ 2.11355 & 1.7592 $\pm$ 0.05631 & 66.1827 $\pm$ 0.22902 \\
NAdam & 0.0558 $\pm$ 0.00040 & 98.3976 $\pm$ 0.05396 & 1.9516 $\pm$ 0.00162 & 70.0233 $\pm$ 0.13590 \\
NAdamW & 0.0479 $\pm$ 0.00101 & 98.7139 $\pm$ 0.01168 & 1.6328 $\pm$ 0.00854 & 71.4420 $\pm$ 0.06245 \\
Nesterov & 0.1572 $\pm$ 0.04028 & 95.4427 $\pm$ 1.31832 & 1.7840 $\pm$ 0.00779 & 66.2933 $\pm$ 0.34269 \\
VeLO & 0.0862 $\pm$ 0.00715 & 97.4058 $\pm$ 0.27273 & 1.5358 $\pm$ 0.02996 & 72.9073 $\pm$ 0.09617 \\
VeLO Short & 0.1445 $\pm$ 0.00158 & 95.6785 $\pm$ 0.04413 & 1.3775 $\pm$ 0.00333 & 73.2160 $\pm$ 0.10806 \\
HPO & 0.5380 $\pm$ 0.00695 & 92.0088 $\pm$ 0.23923 & 1.1170 $\pm$ 0.00383 & 77.4887 $\pm$ 0.11420 \\
\hline
\end{tabular}
\endgroup

\subsection{FastMRI}
\begingroup
\begin{tabular}{lcccc}
\hline
Name & train/loss & train/ssim & validation/loss & validation/ssim \\
\hline
Adam & 0.2702 $\pm$ 0.00693 & 74.2444 $\pm$ 0.51236 & 0.2850 $\pm$ 0.00002 & 72.6139 $\pm$ 0.01243 \\
Heavy Ball & 0.2806 $\pm$ 0.00298 & 72.8935 $\pm$ 0.20797 & 0.2897 $\pm$ 0.00014 & 71.9518 $\pm$ 0.05269 \\
NAdam & 0.2692 $\pm$ 0.00396 & 74.3033 $\pm$ 0.50080 & 0.2851 $\pm$ 0.00022 & 72.6006 $\pm$ 0.03608 \\
NAdamW & 0.2750 $\pm$ 0.00102 & 73.4651 $\pm$ 0.09084 & 0.2851 $\pm$ 0.00015 & 72.5916 $\pm$ 0.04572 \\
Nesterov & 0.2809 $\pm$ 0.00508 & 72.9652 $\pm$ 0.48978 & 0.2898 $\pm$ 0.00005 & 71.9132 $\pm$ 0.02163 \\
VeLO & 0.2737 $\pm$ 0.00324 & 74.0819 $\pm$ 0.41933 & 0.2851 $\pm$ 0.00008 & 72.6663 $\pm$ 0.00503 \\
VeLO Short & 0.2763 $\pm$ 0.00128 & 73.6923 $\pm$ 0.09914 & 0.2850 $\pm$ 0.00026 & 72.6646 $\pm$ 0.02622 \\
HPO & 0.2716 $\pm$ 0.00371 & 74.2704 $\pm$ 0.30959 & 0.2851 $\pm$ 0.00067 & 72.6110 $\pm$ 0.13964 \\
\hline
\end{tabular}

\endgroup

\subsection{Criteo-1TB}
\begin{tabular}{lcc}
\hline
Name & train/loss & validation/loss \\
\hline
Adam & 0.1222 $\pm$ 0.00098 & 0.1237 $\pm$ 0.00005 \\
Heavy Ball & 0.1268 $\pm$ 0.00110 & 0.1279 $\pm$ 0.00094 \\
NAdam & 0.1237 $\pm$ 0.00280 & 0.1255 $\pm$ 0.00317 \\
NAdamW & 0.1226 $\pm$ 0.00059 & 0.1237 $\pm$ 0.00003 \\
Nesterov & 0.1296 $\pm$ 0.00173 & 0.1298 $\pm$ 0.00153 \\
VeLO & 0.1232 $\pm$ 0.00039 & 0.1240 $\pm$ 0.00005 \\
VeLO Short & 0.1236 $\pm$ 0.00024 & 0.1242 $\pm$ 0.00003 \\
HPO & 0.1219 $\pm$ 0.00085 & 0.1237 $\pm$ 0.00012 \\
\hline
\end{tabular}

\subsection{OGBG}
Please note that on HPO results, 2 trials out of 3 were unstable, hence, the missing standard deviations.
\begin{tabular}{lcccc}
\hline
Name & train/loss & train/mAP & validation/loss & validation/mAP \\
\hline
Adam & 0.0165 $\pm$ 0.00045 & 76.3886 $\pm$ 1.51418 & 0.0515 $\pm$ 0.00021 & 27.3651 $\pm$ 0.18156 \\
Heavy Ball & 0.0329 $\pm$ 0.00012 & 31.9587 $\pm$ 0.42343 & 0.0461 $\pm$ 0.00022 & 23.0116 $\pm$ 0.01945 \\
NAdam & 0.0174 $\pm$ 0.00026 & 74.3218 $\pm$ 1.07672 & 0.0509 $\pm$ 0.00027 & 27.2395 $\pm$ 0.63613 \\
NAdamW & 0.0196 $\pm$ 0.00181 & 68.3208 $\pm$ 5.45994 & 0.0483 $\pm$ 0.00157 & 27.6925 $\pm$ 0.23413 \\
Nesterov & 0.0323 $\pm$ 0.00040 & 33.1559 $\pm$ 0.62190 & 0.0458 $\pm$ 0.00024 & 23.6963 $\pm$ 0.38710 \\
VeLO & 0.0164 $\pm$ 0.00037 & 76.6425 $\pm$ 0.92113 & 0.0510 $\pm$ 0.00017 & 27.4374 $\pm$ 0.30907 \\
VeLO Short & 0.0180 $\pm$ 0.00058 & 72.5836 $\pm$ 0.96198 & 0.0491 $\pm$ 0.00064 & 28.2645 $\pm$ 0.35253 \\
HPO & 0.0205 $\pm$ nan & 58.8443 $\pm$ nan & 0.0463 $\pm$ nan & 28.9687 $\pm$ nan \\
\hline
\end{tabular}

\section{Step-Controlled Experiments Results}\label{app:step_controlled}
\subsection{Steps-To-Result Measurements}
\begin{table}[h]
    \centering
    \caption{Number of steps to achieve the train target performance. These numbers are used to plot the performance profiles in figure \ref{fig:step_train_profile}. A value of inf indicates that that baseline was unable to achieve the target within the maximum allowed runtime.}
    \begin{tabular}{lcccc}
        \toprule
        Workload & Criteo & FastMRI & ImageNet & OGBG \\
        Optimizer &  &  &  &  \\
        \midrule
        Adam & \textbf{3497} & \textbf{21262} & \textbf{102029} & \textbf{47419} \\
        HeavyBall & inf & inf & inf & inf \\
        NAdam & 4589 & inf & 121571 & inf \\
        NAdamW & 4694 & inf & 117460 & inf \\
        Nesterov & inf & inf & inf & inf \\
        VeLO & 5252 & inf & inf & 71649 \\
        \midrule
        VeLO Short & 4057 & inf & inf & inf \\
        \bottomrule
    \end{tabular}
\end{table}

\begin{figure}[h]
    \centering
    \caption{Steps-to-Target performance profiles of baselines vs VeLO and VeLO with 75\% (VeLO Short) prompt on training targets.}
    \label{fig:step_train_profile}
    \includegraphics[width=\linewidth]{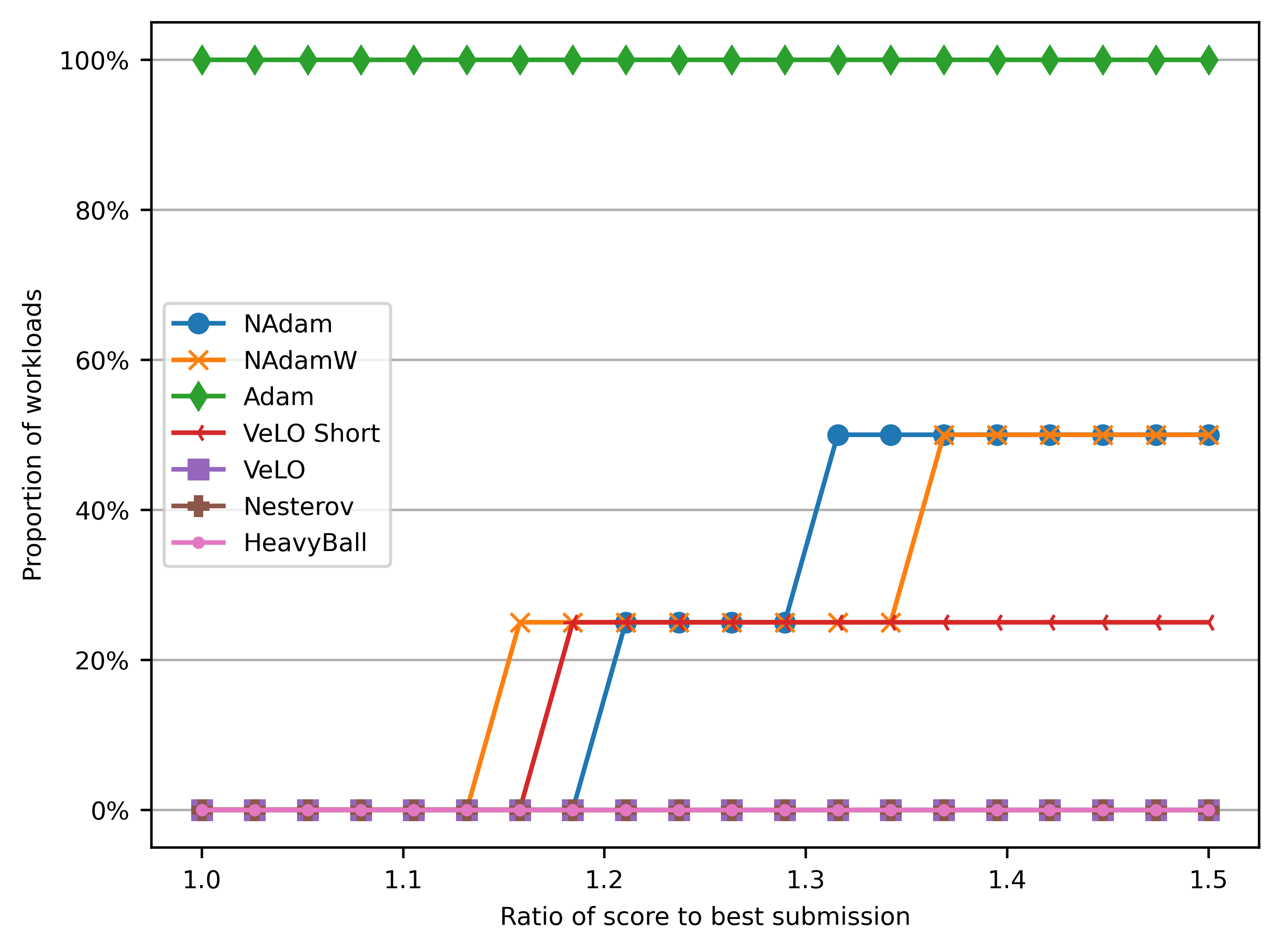}
\end{figure}

\begin{table}[h]
    \centering
    \caption{Number of steps to achieve the validation target performance. These numbers are used to plot the performance profiles in figure \ref{fig:step_val_profile}. A value of inf indicates that that baseline was unable to achieve the target within the maximum allowed runtime.}
    \begin{tabular}{lcccc}
        \toprule
        Workload & Criteo & FastMRI & ImageNet & OGBG \\
        Optimizer &  &  &  &  \\
        \midrule
        Adam & \textbf{5025} & \textbf{13344} & inf & inf \\
        HeavyBall & inf & inf & inf & inf \\
        NAdam & 5177 & inf & inf & inf \\
        NAdamW & 5277 & inf & \textbf{112463} & inf \\
        Nesterov & inf & inf & inf & inf \\
        VeLO & 6105 & 17649 & 114586 & 46979 \\
        \midrule
        VeLO Short & inf & 17616 & 89419 & 34078 \\
        \bottomrule
        \end{tabular}
\end{table}

\begin{figure}[h]
    \centering
    \caption{Steps-to-Target performance profiles of baselines vs VeLO and VeLO with 75\% (VeLO Short) prompt on validation targets.}
    \label{fig:step_val_profile}
    \includegraphics[width=\linewidth]{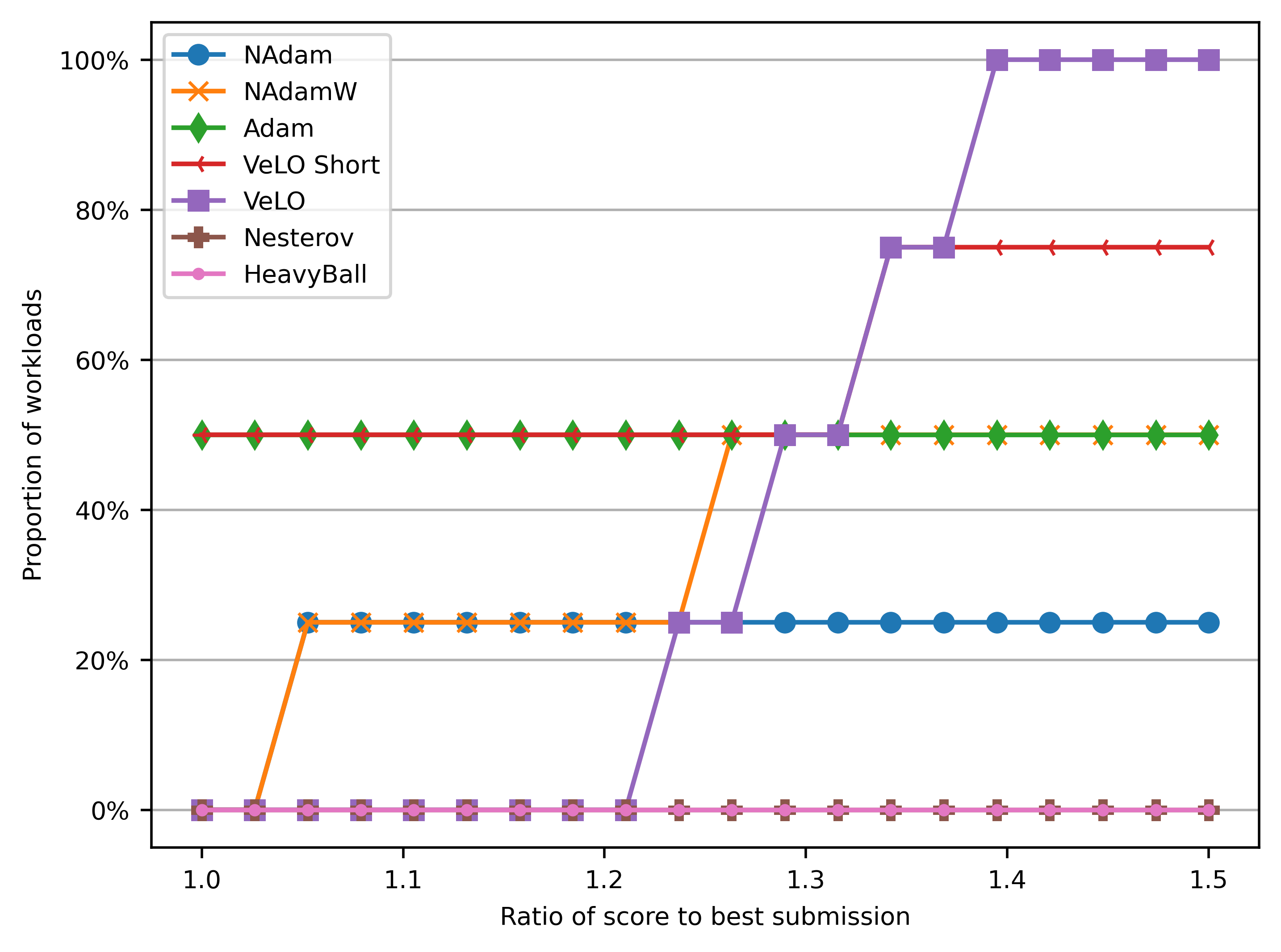}
\end{figure}

\subsection{ImageNet}
\begin{tabular}{lcccc}
\toprule
Name & train/loss & train/accuracy & validation/loss & validation/accuracy \\
\midrule
Adam & 0.0470 $\pm$ 0.00587 & 98.6554 $\pm$ 0.22522 & 1.9438 $\pm$ 0.00717 & 69.8487 $\pm$ 0.05294 \\
Heavy Ball & 0.2582 $\pm$ 0.02502 & 92.2350 $\pm$ 0.80171 & 1.6870 $\pm$ 0.02763 & 66.2940 $\pm$ 0.31674 \\
NAdam & 0.0571 $\pm$ 0.00060 & 98.3279 $\pm$ 0.04514 & 1.9528 $\pm$ 0.00120 & 69.9993 $\pm$ 0.11780 \\
NAdamW & 0.0470 $\pm$ 0.00106 & 98.7786 $\pm$ 0.04481 & 1.6345 $\pm$ 0.01098 & 71.4537 $\pm$ 0.07961 \\
Nesterov & 0.2512 $\pm$ 0.02606 & 92.4685 $\pm$ 0.86639 & 1.7022 $\pm$ 0.03822 & 66.3390 $\pm$ 0.35794 \\
VeLO & 0.1046 $\pm$ 0.00342 & 96.8478 $\pm$ 0.11937 & 1.5017 $\pm$ 0.02693 & 72.9120 $\pm$ 0.09714 \\
HPO & 0.5542 $\pm$ 0.00117 & 91.5016 $\pm$ 0.06893 & 1.1154 $\pm$ 0.00098 & 77.4423 $\pm$ 0.08693 \\
\bottomrule
\end{tabular}

\subsection{FastMRI}
\begin{tabular}{lcccc}
\toprule
Name & train/loss & train/ssim & validation/loss & validation/ssim \\
\midrule
Adam & 0.2702 $\pm$ 0.00693 & 74.2438 $\pm$ 0.51237 & 0.2850 $\pm$ 0.00003 & 72.6131 $\pm$ 0.01279 \\
Heavy Ball & 0.2809 $\pm$ 0.00282 & 72.8451 $\pm$ 0.14562 & 0.2899 $\pm$ 0.00008 & 71.9194 $\pm$ 0.02081 \\
NAdam & 0.2692 $\pm$ 0.00396 & 74.3011 $\pm$ 0.49641 & 0.2851 $\pm$ 0.00022 & 72.5980 $\pm$ 0.04064 \\
NAdamW & 0.2750 $\pm$ 0.00102 & 73.4640 $\pm$ 0.09133 & 0.2851 $\pm$ 0.00016 & 72.5903 $\pm$ 0.04674 \\
Nesterov & 0.2811 $\pm$ 0.00505 & 72.9825 $\pm$ 0.45841 & 0.2899 $\pm$ 0.00004 & 71.9274 $\pm$ 0.00075 \\
VeLO & 0.2804 $\pm$ 0.00008 & 73.5760 $\pm$ 0.01010 & 0.2851 $\pm$ 0.00022 & 72.6630 $\pm$ 0.02184 \\
HPO & 0.2716 $\pm$ 0.00381 & 74.2566 $\pm$ 0.29555 & 0.2850 $\pm$ 0.00080 & 72.6028 $\pm$ 0.13814 \\
\bottomrule
\end{tabular}

\subsection{Criteo}
\begin{tabular}{lcc}
\toprule
Name & train/loss & validation/loss \\
\midrule
Adam & 0.1225 $\pm$ 0.00015 & 0.1237 $\pm$ 0.00005 \\
Heavy Ball & 0.1293 $\pm$ 0.00045 & 0.1299 $\pm$ 0.00062 \\
NAdam & 0.1239 $\pm$ 0.00302 & 0.1256 $\pm$ 0.00315 \\
NAdamW & 0.1220 $\pm$ 0.00018 & 0.1237 $\pm$ 0.00005 \\
Nesterov & 0.1301 $\pm$ 0.00096 & 0.1305 $\pm$ 0.00067 \\
VeLO & 0.1229 $\pm$ 0.00034 & 0.1240 $\pm$ 0.00008 \\
HPO & 0.1222 $\pm$ 0.00058 & 0.1238 $\pm$ 0.00022 \\
\bottomrule
\end{tabular}

\subsection{OGBG}
\begin{tabular}{lcccc}
\toprule
Name & train/loss & train/mAP & validation/loss & validation/mAP \\
\midrule
Adam & 0.0165 $\pm$ 0.00033 & 76.2472 $\pm$ 0.74703 & 0.0515 $\pm$ 0.00020 & 27.3603 $\pm$ 0.18697 \\
Heavy Ball & 0.0341 $\pm$ 0.00026 & 29.8791 $\pm$ 0.24316 & 0.0466 $\pm$ 0.00038 & 22.3993 $\pm$ 0.03845 \\
NAdam & 0.0173 $\pm$ 0.00033 & 74.3650 $\pm$ 0.94344 & 0.0509 $\pm$ 0.00027 & 27.2383 $\pm$ 0.63573 \\
NAdamW & 0.0197 $\pm$ 0.00196 & 68.1042 $\pm$ 5.87819 & 0.0483 $\pm$ 0.00157 & 27.6927 $\pm$ 0.23119 \\
Nesterov & 0.0331 $\pm$ 0.00008 & 31.8606 $\pm$ 0.30560 & 0.0462 $\pm$ 0.00027 & 22.9570 $\pm$ 0.22694 \\
VeLO & 0.0153 $\pm$ 0.00088 & 79.4321 $\pm$ 1.70572 & 0.0522 $\pm$ 0.00126 & 27.5886 $\pm$ 0.47789 \\
HPO & 4.1961 $\pm$ 3.61664 & 61.3348 $\pm$ nan & 4.3632 $\pm$ 3.73861 & 28.9854 $\pm$ nan \\
\bottomrule
\end{tabular}

\newpage
\section{Current vs Original VeLO Evaluation Protocol}\label{app:eval_compar}
The original VeLO evaluation codebase was not released. Nevertheless, the authors do evaluate on a specific version of the MLCommons benchmark that precedes the latest one used in this study \cite{dahl2023benchmarking}. The workloads they chose were:
\begin{enumerate}
    \item Resnet-50 on ImageNet
    \item ViT on Imagenet
    \item Transformer on WMT
    \item Deepspeech on Librespeech
    \item Conformer on Librespeech
    \item GNN on OGBG
\end{enumerate}

Meanwhile, we choose the workloads to be as close as possible to the training distribution:
\begin{enumerate}
    \item Resnet-50 on ImageNet
    \item U-Net on FastMRI
    \item GNN on OGBG
    \item DLRM on Criteo-1TB
\end{enumerate}
Please note that ResNets and Autoencoders (U-Net like architectures) were seen heavily during training. Moreover, GNNs and DLRMs are implemented as MLPs which are also seen heavily during training. ImageNet was part of the training datasets employed too. It is unclear whether the magnitude, sparsity and other properties of the gradients on the above workloads deviate heavily from the totality of meta-training tasks sampled.

Beside the choice of workloads, the original paper have trained (1) VeLO and (2) 20 hyperparameter trials of Adam for a fixed number of steps. Results are presented as figures of training curves which paints comparisons imprecise. Meanwhile, our evaluation uses the MLCommons recommended total wall-clock time to run for and an approximately equivalent total number of steps across a wider set of optimizers. We also fix the optimizers hyerparameters to default values chosen one-shot by the authors to stay faithful to the self-tuning regime instead of sampling for a hyperparameter search space tuned to target validation performance. To paint a more detailed picture and stronger conclusions, we report all final results across step and wall-clock time quotas, and detailed raw data for performance profiles. All in all, our protocol shows that VeLO's efficacy is invalidated once the results are stress-tested in a fair setup.
\end{document}